\pdfoutput=1
\documentclass[sigconf,screen]{acmart}

\usepackage{times}
\usepackage{epsfig}
\usepackage{graphicx}
\usepackage{amsmath}
\usepackage{booktabs}
\usepackage{pgfplots}
\usepackage{multicol}
\usepackage{multirow}
\usepackage{subfigure}
\usepackage{tikz}
\usepackage{appendix}
\usepackage{comment}
\usepackage{emptypage}

\usepackage{fancyhdr}
\renewcommand\footnotetextcopyrightpermission[1]{}

\usepackage[small,compact]{titlesec}
\pgfplotsset{compat=1.17} 
\begin{document}

\setlength{\abovedisplayskip}{1pt} 
\setlength{\belowdisplayskip}{1pt}
\setlength{\floatsep}{2pt plus 2pt minus 2pt}
\setlength{\textfloatsep}{2pt plus 2pt minus 2pt}
\setlength{\intextsep}{2pt plus 2pt minus 2pt}
\title{Generic Attention-model Explainability by Weighted Relevance Accumulation}
\author{Yiming Huang}
\affiliation{%
  \institution{Beijing University of Technology}
  \city{Beijing}
  \country{China}
  }
\email{huangyiming2002@126.com}

\author{Aozhe Jia}
\affiliation{%
  \institution{Beijing University of Technology}
  \city{Beijing}
  \country{China}
  }
\email{jiaaozhe@emails.bjut.edu.cn}

\author{Xiaodan Zhang}
\authornote{Corresponding author}
\affiliation{%
  \institution{Beijing University of Technology}
  \city{Beijing}
  \country{China}
  }
\email{zhangxiaodan@bjut.edu.cn}

\author{Jiawei Zhang}
\affiliation{%
  \institution{Sensetime Research}
  \city{Beijing}
  \country{China}
    }
\email{zhjw1988@gmail.com}

\begin{abstract}
Attention-based transformer models have achieved remarkable progress in multi-modal tasks, such as visual question answering. 
The explainability of attention-based methods has recently attracted wide interest as it can explain the inner changes of attention tokens by accumulating relevancy across attention layers.
Current methods simply update relevancy by equally accumulating the token relevancy before and after the attention processes. 
However, the importance of token values is usually different during relevance accumulation.
In this paper, we propose a weighted relevancy strategy, which takes the importance of token values into consideration, to reduce distortion when equally accumulating relevance.
To evaluate our method, we propose a unified CLIP-based two-stage model, named CLIPmapper, to process Vision-and-Language tasks through CLIP encoder and a following mapper.
CLIPmapper consists of self-attention, cross-attention, single-modality, and cross-modality attention, thus it is more suitable for evaluating our generic explainability method. 
Extensive perturbation tests on visual question answering and image captioning validate that our explainability method outperforms existing methods.
\end{abstract}



\maketitle

\section{Introduction}
In recent years, multi-modal learning has achieved rapid progress in Vision-and-Language (V+L) tasks like image captioning~\cite{caption} and visual question answering~\cite{VQAv2}. The milestone work of CLIP~\cite{CLIP} marked that large-scale attention-based Transformer architecture has become mainstream to learn richer features in pretraining and perform better in downstream V+L tasks. Meanwhile, researchers who focus on explainable AI gradually carry out their study in multi-modal situations where inspiring works give explanations from different views.

\begin{figure}[!htbp]
	\begin{center}
		\includegraphics[width=0.7\linewidth]{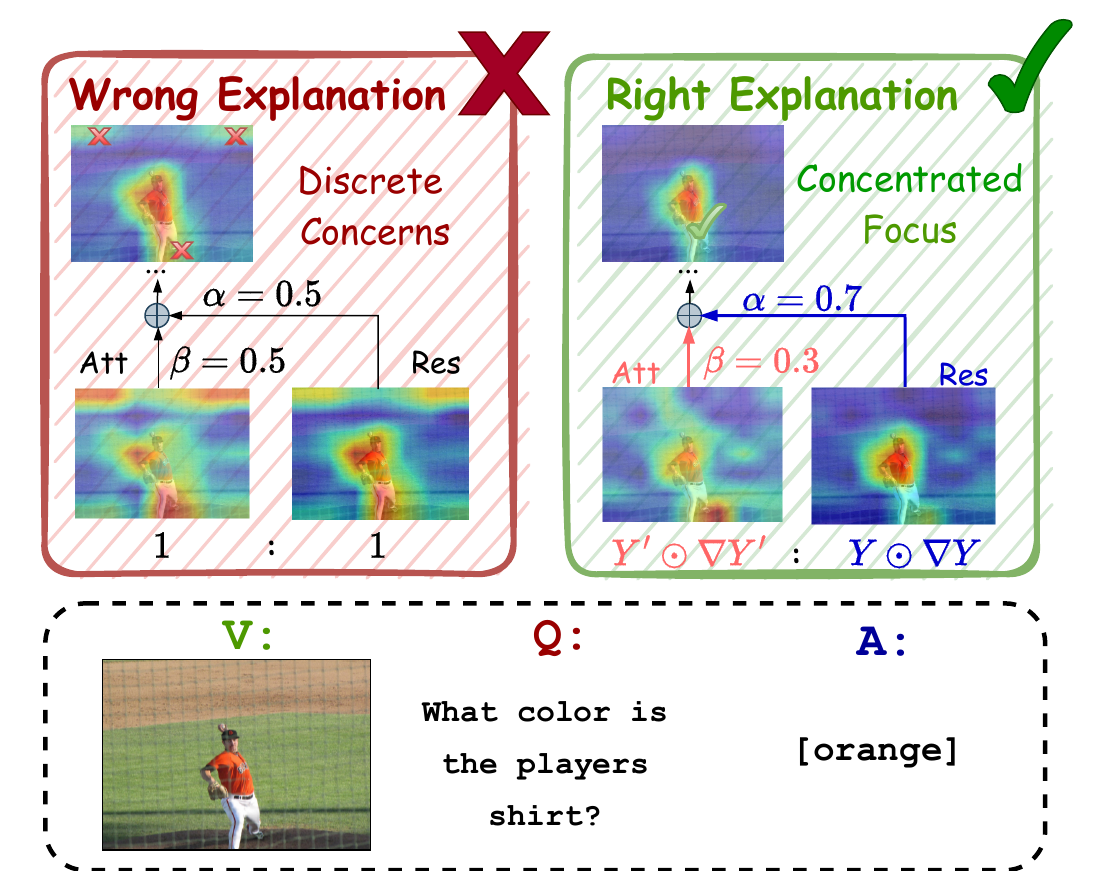}
	\end{center}
	\caption{An illustration of our motivation. We compare our method with the ablation version of our method that regards the same importance of relevance before attention and relevance after residual connection. We visualize layer 4 of CLIP-ViT encoder (our CLIPmapper encoder) at the inference period of VQA task, ``Res" is the visualization of relevance in residual connection and ``Att" is the visualization of relevance in attention.}
	\label{fig:motv}
\end{figure}

Attention Rollout~\cite{Rollout} distinctly points out that attention is the process of token-linear-combination and recursively multiplies the attention weights in all the layers to establish the relevancy between tokens, but it only simply averages importance and relevance across heads. GenAtt~\cite{GenAtt} further considers the equivalent attention across heads to avoid distorted relevance by introducing AttGrad \\ ~\cite{rethink} to balance attention gradient and values in relevance computation.\label{sec:g1}
However, these methods equally take into account integrating the relevance in attention and residual connection, which might fail to restore the significant changes of attention. As shown in Fig.~\ref{fig:motv}, under the equal integration of residual connection and attention, the relevancy map tends to focus on discrete areas, which seriously damages the explainability of attention.


In this paper, we propose a novel relevancy-weighted method for improving generic attention-oriented explainability. In relevancy aggregation, our method adaptively adjusts weights for attention and residual connection according to the importance of tokens, which reduces distortion from rigid average weighting. As shown in Fig.~\ref{fig:motv}, our weighted relevancy is able to adaptively assign ratios for attention and residual connection in relevancy integration, resulting in more focused attention areas.
To evaluate our method, we propose a unified CLIP-based two-stage model, named CLIPmapper, to fulfill different tasks through attention-based CLIP encoder and mappers. CLIPmapper consists of self-attention, cross-attention, single-modality, and cross-modality attention, which covers most of the attention types in multi-modal learning. The experimental results on visual question answering and image captioning validate that our method extends the generic attention explainability. 

The main contributions of this paper are demonstrated as follows: (1) We propose a novel attention explainability method by introducing an adaptively weighted strategy for relevancy aggregating. (2) We propose a unified vision and language model, CLIPmapper, which covers most of the attention types in multi-modal learning for testing explanation methods roundly. (3) We extend our method to show modality interaction and to explain the generative task. The perturbation tests and visualization results on VQA and image captioning show an improvement in explainability.

\section{Related Works}

\subsection{Design and Analysis of V+L Models}
The current mainstream V+L large models can be divided into two architectures: single-stream and dual-stream, which differently learn the joint multi-modal representation with self-attention and coordinated multi-modal representation with cross-attention ~\cite{VLsurvey1}~\cite{VLsurvey2}.

The single-stream model is stacked by Transformer blocks based on self-attention. Its input is usually the concatenation of image tokens and text tokens, like ViLT~\cite{ViLT}, VisualBert~\cite{VisualBert}, Oscar~\cite{Oscar}, etc. They implicitly implement the fusion of modalities through the general computing ability of the transformer. Thereby, representing the two modalities in a joint representation space, which is the idea behind our baseline model is Mapper A.

In the dual-stream model, each modality has a corresponding input and output. They generally use contrastive learning to align modalities and use cross-attention for interactions between modalities, such as LXMERT~\cite{LXMERT}, ALBEF~\cite{ALBEF}, Coca~\cite{Coca}, etc. Our Mapper B follows this idea.

We also noticed many analytical works on these V+L models. For these two architectures, VOLTA\cite{VOLTA} has shown that their performance of the downstream task is almost the same with the same number of parameters. Our experiments also confirmed this. CLIP-ViL~\cite{CLIP-ViL} and ~\cite{v4l} have provided qualitative conclusions from the perspective of performance experiments, while our work tends to be explanatory. Works such as~\cite{MNER} and~\cite{MProbe} tend to judge inherent properties of V+L models through subtle-designed experiments, while our work focuses on quantifiable and visualizable methods.

\subsection{Explainability Methods of Multi-modal Models}
There are not sufficient general studies like GenAtt about the explainabilities in multi-modal models and we have found that existing work has various limitations. The first is the explanation-oriented model design, like~\cite{careful1} and~\cite{careful2}. Despite careful design making explainability, it is hard to be generalized for mainstream works. On the contrary, our method aims to take effect generally on V+L Transformers. The second is the restricted task limitation, like~\cite{task1} and~\cite{task2}, it is also hard to generalize, too. While DIME~\cite{DIME} attracts our attention, it inherits the model-agnostic trait of LIME and gives the explanation for multi-modal interactions. Although our method focuses on different concerns in this problem that our explanation is more faithful to attention models.

\subsection{Explainability Methods of Attention-model}
Debates about whether attention can be explained~\cite{isntEXP}~\cite{isEXP} have lasted for a long period. GenAtt performs well in perturbation tests~\cite{rethink}, we attribute this outperformance to the effect of attention gradients $\mathbf{\odot}$ attention values method (AttGrad, $\mathbf{\odot}$ is element-wise production). We believe this method is effective because gradients can reflect the importance of neurons to the final output of the model when the activation value is small, while the activation value itself reflects importance when it is large. Therefore, multiplying them together can represent neuron importance well. GenAtt gets equivalent attention across heads by weighting attention with this method, which is the reason why GenAtt beat Rollout whose equivalent attention is only the simple average across heads.\label{sec:g2} But on the issue of balancing the different effects of attention and residual connection, they both simply add their relevancy scores to the scores after attention to embody the effects of residual connection. However, we hold the view that we can do the balancing with the same method element-wise multiplying values and gradients because it reflects the significance of the final output of the model.


\section{Methodology}
\subsection{Weighting Residual Connection for Attention Explaination}
\begin{figure}[!hbtp]
	\begin{center}
		\includegraphics[width=0.7\linewidth]{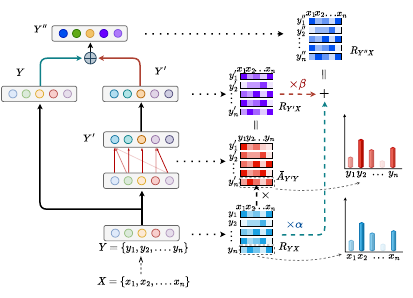}
	\end{center}
	\caption{Illustration of relevance change of tokens process in attention block. $\mathbf{Y = \{y_1, y_2,...y_n\}}$ is the input of this layer that we know its relevance of $\mathbf{X = \{x_1, x_2,...x_n\}}$ (the original input of the whole model) before this attention block. It turns to $\mathbf{Y' = \{y'_1, y'_2,...y'_n\}}$ after attention. Finally, it becomes the $\mathbf{Y''}$ due to residual connection. The red and green lines represent different effects between attention and residual connection, corresponding to $\mathbf{\alpha}$ and $\mathbf{\beta}$. All the first index of the map means tokens and the second means its relevance about corresponding previous tokens.}
	\label{fig:exp}
\end{figure}
\subsubsection{Explainability in Simplified Self-Attention Case.} It is obvious that input tokens have clear semantics, such as the patch of images or the subword of text. As long as we determine the reasonable and quantitative relationship between input and output tokens, we can obtain an explanation that is intuitive for comprehension.

As pointed out in the work of Rollout, the attention operation itself is the linear recombination of tokens. We noticed that only the attention operation and residual connection in the vanilla Transformer model change the relevance of the output tokens.

This change in relevance is easy to trace layer-wise from the bottom up of the model. As GenAtt inspired us, we fairly measure the influence brought by attention and residual connection by weighting with gradient $\mathbf{\odot}$ activation values.

Considering the simplified case of a single self-attention as Fig.~\ref{fig:exp} displayed for the multi-modal classification task like VQA, we assume that the tokens $\mathbf{X = \{x_1, x_2,...x_n\}}$ are the original input before entering the model and we know the relevance accumulated by previous layers of current layer input tokens  $\mathbf{Y = \{y_1, y_2,...y_n\}}$, represented by relevance map $\mathbf{R}$. Each row in $\mathbf{R}$ represents how each token in Y is relevant to different proportions about each token in X, and the sum is 100\%. When $\mathbf{Y}$ is transformed into $\mathbf{Y'}$ after the Attention operation, the corresponding relevance map also changes to $\mathbf{R'}$. This change can be expressed as follows:
\begin{align}
    &\label{eq1}\mathbf{R_{Y'X} = \bar{A}_{Y'Y} \cdot R_{YX}}\\
    &\label{eq2}\mathbf{\bar{A} = norm_{row}(E_h((\nabla A \odot A)^{+}))}\\
    &\label{eq3}\mathbf{\nabla A = \frac{\partial z_t}{\partial A}}.
\end{align}
Here, to ensure the sum is 100\%, we use $\mathbf{norm_{row}}$ normalize each row of the equivalent attention, $\mathbf{E_h}$ is the average over the heads, $\mathbf{\odot}$ is element-wise multiplication, $\mathbf{(.)^{+}}$ means we only reserve positive relevance, $\mathbf{z_t}$ corresponds to the prediction of the model in the multi-modal classification task, and $\mathbf{A}$ is attention scores of
current layer.
As for the residual connection, we argue the different importance of relevance in attention and relevance in residual connection is the vital term for better relevance propagation. On the other hand, as mentioned in Sec.~\ref{sec:g1} and Sec.~\ref{sec:g2} gradients and values can respectively represent the sensitivity and the influence of tokens, we comprehensively consider these two things as important. The abovementioned method element-wise multiplying input values gradients is called InputGrad in~\cite{rethink}, we believe this method can play the same role as AttGrad due to combining value and gradient, which alleviates the ``gradient saturation" phenomenon~\cite{rethink} to express the importance of token (or neuron cell). Therefore, take it as weight is appropriate.

To be specific, the formulas are as follows ($\mathbf{\alpha}$, $\mathbf{\beta}$ is the sensitivity weights, $\mathbf{(.)^{+}}$ means only positive value can be counted as importance, their sum is 100\%, $\mathbf{avg(.)}$ is the scalar average of the whole matrix):
\begin{align}
    &\label{eq4}\mathbf{R_{Y''X} = \alpha \cdot R_{YX} + \beta \cdot R_{Y'X}}\\
    &\label{eq5}\mathbf{\mathbf{\alpha} = avg(\frac{(\nabla Y \odot Y)^{+}}{(\nabla Y \odot Y)^{+} + (\nabla Y' \odot Y')^{+}})}\\
    &\label{eq6}\mathbf{\beta = 1 - \alpha}\\
    &\label{eq7}\mathbf{\nabla Y = \frac{\partial z_t}{\partial Y}}.
\end{align}
\begin{figure}[!hbtp]
	\begin{center}
		\includegraphics[width=0.7\linewidth]{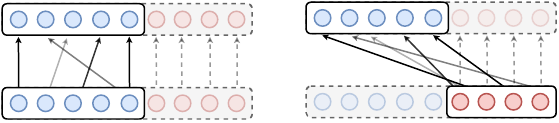}
	\end{center}
	\caption{We can associate unimodal self-attention and cross-attention with another absent modality token. }
	\label{fig:multi}
\end{figure}
\subsubsection{Obtaining Relevancy Map for Bimodal Cases.} Next, we consider the update rule of layerwise changes-tracing of the relevance map in general multi-modal Transformers. First, before going through the Transformer, each token is composed of itself, which is the initialization of our relevance map $\mathbf{R}$. That is:
\begin{align}
    &\label{eq8}\mathbf{R = \mathbb{I}_{(s+q)\times(s+q)}}.
\end{align}
As for the case like our Mapper A in Sec.~\ref{sec:ma} that the input tokens are the concatenation of image tokens and text tokens, this joint modality $S$ can use the Eq.~\ref{eq4} to update relevance, that is:
\begin{align}
    &\label{eq9}\mathbf{R = \alpha \cdot R + \beta \cdot \bar{A}_{S'S} \cdot R}.
\end{align}
For unimodal self-attention of modality $\mathbf{S}$, just imagine the absent tokens of modality $\mathbf{Q}$ do attention operation at the same time with attention scores 1 to itself like the Fig.~\ref{fig:multi} left part shows. Therefore, the update rule is:
\begin{align}
    &\label{eq10}\mathbf{R = \alpha \cdot R + \beta \cdot \begin{vmatrix} 
     	\mathbf{\bar{A}_{S'S}} & 0 \\
     	0 & \mathbb{I}
     \end{vmatrix} \cdot C }.
\end{align}
For multi-modal cross-attention whose query modality is $\mathbf{S}$ and key/value modality is $\mathbf{Q}$, do same imagination as Fig.~\ref{fig:multi} right part shows, the update rule is:
\begin{align}
    &\label{eq11}\mathbf{R = \alpha \cdot R + \beta \cdot \begin{vmatrix} 
     	0 &\mathbf{\bar{A}_{S'Q}} \\
     	0 & \mathbb{I}
     	\end{vmatrix} \cdot R }.
\end{align}
In the above formulas:
\begin{align}
    &\label{eq12}\mathbf{\bar{A} = norm_{row}(E_h((\nabla A \odot A)^{+}))}\\
    &\label{eq13}\mathbf{\mathbf{\alpha} = avg(\frac{\nabla S \odot S}{\nabla S \odot S + \nabla S' \odot S'})}\\
    &\label{eq14}\mathbf{\beta = 1 - \alpha}\\
    &\label{eq15}\mathbf{\nabla A = \frac{\partial z_t}{\partial A}, \nabla S = \frac{\partial z_t}{\partial S}, \nabla S' = \frac{\partial z_t}{\partial S'}}.
\end{align}
That is our unified form of update rules for relevance changes in cases of multi-modal classification tasks.

\subsection{CLIPmapper: a New Baseline Model for Examining Explainability Methods} 
\begin{figure*}[!htbp]
\begin{center}
\includegraphics[width=0.75\linewidth]{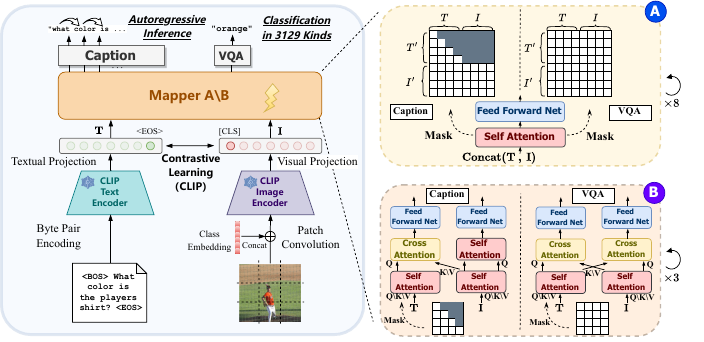}
\end{center}
   \caption{The detailed design of CLIPmapper, which has different variations of mappers in A and B.}
\label{fig:arch}
\end{figure*}

We designed a two-stage model based on CLIP encoders as Fig.~\ref{fig:arch} presents for examining the attention-oriented explainability method. It uses frozen CLIP encoders to extract unimodal features and applies the mapper to generate multi-modal semantics in downstream tasks. In the image encoding stage, CLIP-ViT-32-B~\cite{CLIP}extracts grid features rather than previous work like LXMERT or ViusalBert~\cite{VisualBert} using Faster-RCNN~\cite{fasterRCNN}to extract region features which helps explainability methods pay early attention before playing their role. In the text encoding stage, CLIP text decoder is the autoregressive language model that needs the sequence mask to extract features of subwords, which represents a common case in language transformers. \label{sec:modelarch}
For the mapper stage, we consider the commonly adopted single-stream and dual-stream architectures in V+L tasks. These two mappers respectively correspond to the multi-modal self-attention and the cross-attention for modal interaction. Furthermore, We freeze the parameters of the encoder stage and only train the mapper for lower computational cost.

For the visual input of the model, we reserve the preprocessing of vanilla CLIP Vision Transformer~\cite{ViT} encoder that concatenates class embedding with patch projection of the image. For the language input of the model, we do not fix the length of the input tokens in CLIP for better performance on downstream tasks, but still, use subword embeddings of CLIP based on its byte pair encoding (BPE) vocab.

In the image captioning task, we autoregressively generate candidate sentences. In the VQA task, we treat it as a 3129 classes soft-label~\cite{Tips} classification task like previous works.

\subsubsection{Mapper A: Single-Stream Mapper with Self-Attention.} As Fig.~\ref{fig:arch} (A) demonstrates, concatenating visual and textual tokens and feeding them into transformer blocks stacked with self-attention and the feed-forward layer is a common choice in many models, as it simplifies the structure and relies on the general computing ability of transformer for multi-modal interaction. Our Mapper A adopts this structure.\label{sec:ma}

Our single-stream architecture is stacked with modules consisting of self-attention and following feed-forward layers.

In the caption task, considering the autoregressive language modeling, text requires sequence masking, and the image cannot attend to the language since doing so would leak information from future text, breaking the autoregressive nature of language.

\subsubsection{Mapper B: Dual-Stream Mapper with Cross-Attention.} As for the nature of attention itself, where Query (Q), Key (K), and Value (V) matrices may come from different sources, it is easy to facilitate cross-modal interaction with attention, especially when many works use explicit modality interaction designs with different modalities for Q and K/V. This design allows models to form a dual-stream structure, and our Mapper B is designed in this way, as Fig.~\ref{fig:arch} (B) shows.

Our dual-stream structure is designed with two branches per layer, consisting of self-attention, cross-attention, and feed-forward blocks. The two branches exchange K/V input for multi-modal interaction.

Same as Mapper A, we cancel the Key/Value exchange from the text to the image branch to prohibit the leak of future text in the image captioning task. Moreover, we also make sequence masking in self-attention of the text branch.
\subsection{Visualizable Explanation based on Relevance Map}
\subsubsection{The VQA Task.}
\begin{figure}[!htbp]
	\begin{center}
		\includegraphics[width=0.7\linewidth]{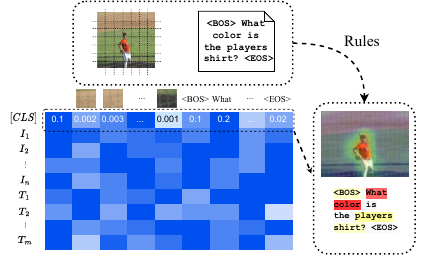}
	\end{center}
	\caption{Visualizing Explanation by relevance Scores of the [CLS] token.}
	\label{fig:vqa}
\end{figure}
For multi-modal classification tasks such as VQA, it is apparent that our explanation is the relevance score of the classification token [CLS], as shown in Fig.~\ref{fig:vqa}.

\subsubsection{The Modality Interaction in VQA.}
\begin{figure}[!htbp]
	\begin{center}
		\includegraphics[width=0.4\linewidth]{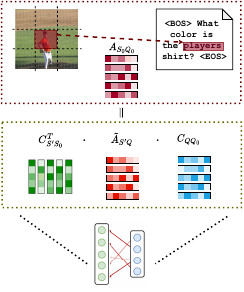}
	\end{center}
	\caption{Demonstration of Modified Interaction Map computed by the relevance of tokens of each modality and their attention.}
	\label{fig:inter}
\end{figure}
In causal works, they all simply take attention to account the interaction between tokens. But according to the definition of our relevance map, the attention of each layer is the interaction of composite tokens which already mix the relevance of other tokens except for the first layer of attention. Thus, we can restore these composite interactions of tokens to the input interaction of tokens. In other words, we restore the attention of these tokens to the attention of the image patches to the subwords (or the opposite) which represent the modality interaction, as shown in Fig.~\ref{fig:inter}. We define it as the Modified Interaction Map $\mathbf{M_{S_0Q_0}}$. Similar to our update rules of the relevance map, we defined the update rules of the Modified Interaction Map as below ($\mathbf{s}$ and $\mathbf{q}$ respectively means token numbers of modalities $\mathbf{S}$ and $\mathbf{Q}$):
\begin{align}
    &\label{eq16}\mathbf{M_{S_0Q_0} := 0_{s\times q}}\\
    &\label{eq17}\mathbf{M_{S_0Q_0} = M_{S_0Q_0} + M_{S'S_{0}}^{T} \cdot \tilde{A}_{S' Q}\cdot M_{Q Q_{0}}}\\
    &\label{eq18}\mathbf{\tilde{A} = E_h((\nabla A \odot A)^{+})}.
\end{align}
The reason why we don't normalize $\mathbf{\tilde{A}}$ in Eq.~\ref{eq18} is to make equivalent attention $\mathbf{\tilde{A}}$ keep the value size of attentions in different layers that embody the different significance of different attentions about the final decision of model. 

Be cautious Eq.~\ref{eq17} is only a one-way interaction which means modality $\mathbf{S}$ pays attention to modality $\mathbf{Q}$. To make it be mutual interaction, we shall consider another one-way interaction. The mutual modified interaction $\mathbf{MM}$ is multiplication as follows:
\begin{align}
    &\label{eq23}\mathbf{MM_{S_0Q_0} = M_{Q_0S_0}^T \odot M_{S_0Q_0}}.
\end{align}
\subsubsection{Explanation of Image Captioning in Word-Level and Sentence-Level.}
To explain the caption task, we can treat it as a classification task in each inference timestep. Therefore, applying our method is feasible. Besides the explanation in each timestep, we can weight them as Eq.~\ref{eq19} to represent the entire explanation for the generation process (Where the whole generated caption is $\mathbf{T}$):
\begin{align}
     &\label{eq19}\mathbf{Score(I) = \frac{\sum^{|T|}_i\alpha_iC_{i(-1,:)}}{\sum^{|T|}_i\alpha_i}}.
\end{align}
Further, we suppose that the importance of generated words (subwords) in each timestep is reflected by how much the evaluation metric drops when it is absent compared to the full caption. we express it as follows\label{sec:eval}:
\begin{align}
     &\label{eq20}\mathbf{\alpha_i = eval(T) - eval(T - {w_i})}.
\end{align}
Fast-Computed metrics like CIDEr~\cite{CIDEr}, METEOR~\cite{METEOR} is suitable to be the $\mathbf{eval(.)}$.

\section{Experiments}
\subsection{Training Setup of CLIPmapper}
\subsubsection{VQA Setup.}
We choose VQAv2~\cite{VQAv2} task to carry out our experiment. It asks for answers to the given image question pairs. 


We train our CLIPmapper(with frozen CLIP encoders) on the train and validation sets while reserving 10000 examples from the validation set to report the true generalization ability of the model. In training, we adopt Adam optimizer with the learning rate $5e-4$ and we choose $64$ as the batch size. The warmup strategy is applied in the first 1000 steps and we use a cosine schedule decaying strategy to reduce the learning rate after 10000 steps.

For a fair comparison between mapper A and mapper B,  we set mapper A to have 8 layers and mapper B to have 3 layers so that the number of parameters of both mappers is almost equal.
\subsubsection{Image Captioning Setup.}

Regarding the image captioning task, we select MSCOCO 2014 caption task~\cite{caption} for the experiment. It requires our model to generate the caption for the given image.

As same as previous studies, we train our CLIPmapper(with frozen CLIP encoders) in Karpathy train-split~\cite{kar} and we regard its performance in karpathy val-split as its generalization ability. For training, we adopt Adam optimizer with the learning rate $1e-5$ and we choose $64$ as the batch size. The warmup strategy is applied in the first 1000 steps. \label{sec:details}

We also set mapper A to have 8 layers and mapper B to have 3 layers for a fair comparison. \label{sec:hyperpara}

\subsection{Performance of CLIPmapper}
\begin{table}[!htbp]  
\caption{Comparison of CLIPmapper with other VQA models. As for TAP-C~\cite{TAP}, they use CLIP-ViT-16-B (*) as encoder. Flamingo are the same low computational cost training model. }
\begin{center} 
\Large
\centering
\tabcolsep=1pt
\scalebox{0.675}{
\begin{tabular}{cc cccc cccc}
\toprule
\multirow{2}{*}{Model} & \multicolumn{4}{c}{test-dev} & \multicolumn{4}{c}{test-std}\\ \cmidrule(lr){2-5} \cmidrule(lr){6-9}
& yes/no & number & others &  \textbf{overall} &  yes/no & number & others &  \textbf{overall} \\ \midrule
CLIPmapper-A (ours) & 81.44 & 44.29 & 54.78 & 64.57 & 81.74 & 44.78 & 55.22 & 65.08\\
CLIPmapper-B (ours) & 81.93 & 44.4 & 56.11 & 65.42 & 82.57 & 44.33 & 56.36 & 65.91\\
BUTD & - & - & - & - & 86.60 & 48.64 & 61.15 & 70.34\\
Flamingo (32 shots) & - & - & - & 67.6 & - & - & - & -\\
TAP-C (32 shots*) & 73.60 & 32.55 & 35.02 & 49.19 & - & - & - & -\\
VisualBert & - & - & - & 70.80  & - & - & - & 71.00\\
LXMERT & - & - & - & 72.4 & 88.2 & 54.2 & 63.1 & 72.5\\
\bottomrule

\end{tabular}  
}
\end{center}
\label{tab:vqa} 
\end{table}

\begin{table}[!htbp]  
\caption{Comparison of CLIPmapper with other image captioning models. B@4 is BLEU-4, C is CIDEr, M is METEOR, and S is SPICE. Clipcap uses its transformer mapping network (**) and Flamingo are the same low-cost training model. BUTD is the typical baseline of the image captioning model. 
To be consistent with explaining the image captioning task, we only do the greedy search (*) to generate captions.}  
\begin{center} 
\Large
\centering
\tabcolsep=1pt
\scalebox{0.7}{
\begin{tabular}{ccccc}
\toprule
Model & B@4 & C & M & S \\ \midrule
CLIPmapper-A (ours*) & 31.8 & 105.9 & 26.6 & 19.9 \\
CLIPmapper-B (ours*) & 31.9 & 108.0 & 26.9 & 20.2\\
BUTD & 36.2 & 113.5 & 27.0 & 20.3 \\
Flamingo (32shots) & - & 113.8 & - & - \\
Clipcap (**) & 33.5 & 113.0 & 30.4 & 23.1 \\
\bottomrule
\end{tabular}  
}
\end{center}
\label{tab:cap} 
\end{table}
We evaluate our CLIPmapper on VQA test-std dataset, VQA test-dev dataset~\cite{VQAv2}, and Karpathy test-split~\cite{kar} of COCO caption dataset. The results compared to other models are shown in Tab.~\ref{tab:vqa} and Tab.~\ref{tab:cap}. These results indicated that (1) our CLIPmapper is qualified to be the examining model because it has no excessive performance gap compared to LXMERT and VisualBert (2) our CLIPmapper achieved considerable generalization ability relative to low-cost training models like Flamingo~\cite{Flamingo}, TAP-C~\cite{TAP}, Clipcap~\cite{ClipCap} (3) our CLIPmapper has the close performance (to be consistent with explaining the image captioning task, we only do the greedy search to generate captions) to the typical baseline BUTD~\cite{BUTD} hence CLIPmapper can represent the general V+L model.


\subsection{Validation for Our Explainability Method}
\subsubsection{Baseline Methods for Comparison:} We are concerned about the explainability of attention models. The current effective methods are Rollout~\cite{Rollout}, TransAtt~\cite{TransAtt}, GradCAM~\cite{gradCAM}, PRLP~\cite{PLRP} and GenAtt~\cite{GenAtt}. Among these methods, GenAtt has state-of-the-art performance and our methods can be seen as an improvement to Rollout so that we validate our method in the comparison with raw attention, Rollout, and GenAtt, \textbf{implementation details can be seen in Appendix}.

\subsubsection{Validating Our Method by Perturbation Tests:} We carry the same perturbation tests as Chefer et al. have done. We gradually mask the corresponding image patch or text subword tokens by the scores derived from these explanatory methods in descending order as the positive perturbation tests. The faster the performance decreases, the more faithful the explanation is, so it can be measured by the area under the curve (AUC) to reflect the faithfulness from the sensitivity side. On the contrary, negative perturbation tests are conducted in ascending order of scores. The larger the area under the curve is, the more faithful in robustness side is. We also choose the same descending/ascending steps as Chefer et al. That is masking in the order of the top 0\%, 5\%, 10\%, 15\%, 20\%, 25\%, 50\%, 75\%, and 100\% high-scoring/low-scoring tokens.

\subsubsection{Ablation Study: Does Weighting Relevance in Residual Connection and Attention Works?} In this part, We take $\mathbf{\alpha = \beta = 0.5}$ as the ablation baseline which represents the common practice of averaging relevance before attention and relevance after attention simply. We take an ablation study to test whether the Weighting works using CLIPmapper-A on reserving 10000 examples of VQAv2 local testing. The perturbation 
tests are conducted on texts. The results are as Fig.~\ref{fig:ablation} shows. 
The results clearly show Weighting makes our method more sensitive/robust in positive/negative tests. This proves that our motives are right.
\begin{figure}[!htbp]
    \begin{tikzpicture}[scale=0.4]
        \begin{axis}[
            xlabel=(a)\% of tokens removed,
            ylabel= accuracy,
            tick align=outside,
            legend pos = north east
            ]
        \addplot[mark=*,red] plot coordinates {
            (0, 0.6122700000405311)
(5, 0.6117700000166894)
(10, 0.5080600001096726)
(15, 0.43958999981880187)
(20, 0.383759999614954)
(25, 0.3306699989497662)
(50, 0.19119999969005586)
(75, 0.07932999985218048)
        };
        \addlegendentry{Ours(AUC=0.230)}
        \addplot[mark=*,yellow] plot coordinates {
            (0, 0.6122700000405311)
(5, 0.6119400000214577)
(10, 0.5191000004470349)
(15, 0.45764999997615813)
(20, 0.40921999953389165)
(25, 0.3635999992728233)
(50, 0.19418999953866006)
(75, 0.07590999994874001)
        };
        \addlegendentry{Ablation(0.237)}
    
    \end{axis}
    \end{tikzpicture}
    \begin{tikzpicture}[scale=0.4]
        \begin{axis}[
            xlabel=(b)\% of tokens removed,
            ylabel= accuracy,
            tick align=outside,
            legend pos = north east
            ]
        \addplot[mark=*,red] plot coordinates {
            (0, 0.6122700000405311)
(5, 0.6122700000405311)
(10, 0.612529999935627)
(15, 0.612350000011921)
(20, 0.61283000010252)
(25, 0.612150000333786)
(50, 0.5872100010335446)
(75, 0.48427999951839446)
        };
        \addlegendentry{Ours(AUC=0.498)}
        \addplot[mark=*,yellow] plot coordinates {
            (0, 0.6122700000405311)
(5, 0.6122700000405311)
(10, 0.6120399998664856)
(15, 0.6122100001335145)
(20, 0.610850000333786)
(25, 0.6074800001978874)
(50, 0.5772500005602836)
(75, 0.4579999996304512)
        };
        \addlegendentry{Ablation(0.489)}
    
    \end{axis}
    \end{tikzpicture}
    \caption{The perturbation tests on texts with CLIPmapper-A for ablation study. The left one is positive test, the right one is negative test. Fig.~\ref{fig:perCap} is same.}
    \label{fig:ablation}
\end{figure}
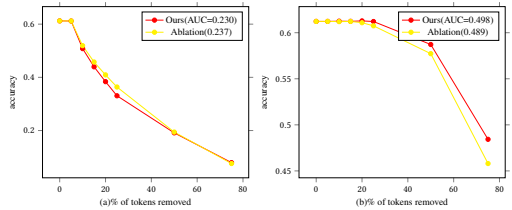

\begin{figure}[!htbp]
    \begin{tikzpicture}[scale=0.4]
        \begin{axis}[
            xlabel=(a)\% of tokens removed,
            ylabel= accuracy,
            tick align=outside,
            legend style={fill=white, fill opacity=0.2, draw opacity=1, text opacity=1},
            legend pos=north east
            ]
        \addplot[mark=*,red] plot coordinates {
            (0, 0.6122700000405311)
(5, 0.5841300000786781)
(10, 0.5620800003409385)
(15, 0.5558400005340576)
(20, 0.5377300001084805)
(25, 0.5323499998807907)
(50, 0.4807599997878075)
(75, 0.4318199989974499)
  
        };
        \addlegendentry{Ours(AUC=0.435)}
        \addplot[mark=*,blue] plot coordinates {
            (0, 0.6122700000405311)
(5, 0.5688400002360344)
(10, 0.5398500007271767)
(15, 0.5269000001370907)
(20, 0.5084999996423721)
(25, 0.49878999970555304)
(50, 0.4514399994790554)
(75, 0.4113499986767769)
  
        };
        \addlegendentry{GenAtt(0.413)}
        \addplot[mark=*,pink] plot coordinates {
           (0, 0.6122700000405311)
(5, 0.581580000257492)
(10, 0.5636600006937981)
(15, 0.5536300008714199)
(20, 0.5371400002777577)
(25, 0.5284200006783009)
(50, 0.4791199996471405)
(75, 0.42913999876379966)
  
        };
        \addlegendentry{Rollout(0.433)}
        \addplot[mark=*,green] plot coordinates {
            (0, 0.6122700000405311)
(5, 0.5805600005149841)
(10, 0.5555800005316734)
(15, 0.5451000008761883)
(20, 0.5303800003707408)
(25, 0.5209700001657009)
(50, 0.46966999974250795)
(75, 0.42769999925494195)
  
        };
        \addlegendentry{RawAtt(0.428)}
    \addlegendentry{line}
    
    \end{axis}
    \end{tikzpicture}
    \begin{tikzpicture}[scale=0.4]
        \begin{axis}[
            xlabel=(b)\% of tokens removed,
            ylabel= accuracy,
            tick align=outside,
            legend style={fill=white, fill opacity=0.2, draw opacity=1, text opacity=1},
            legend pos=south west
            ]
        \addplot[mark=*,red] plot coordinates {
            (0, 0.6122700000405311)
(5, 0.611989999961853)
(10, 0.6109900002360344)
(15, 0.610980000090599)
(20, 0.6100900001525879)
(25, 0.6091200000286102)
(50, 0.6049600000560283)
(75, 0.5693400004982948)
  
        };
        \addlegendentry{Ours(AUC=0.522)}
        \addplot[mark=*,blue] plot coordinates {
            (0, 0.6122700000405311)
(5, 0.6105099999904633)
(10, 0.6106300001263618)
(15, 0.6111200002074242)
(20, 0.6104800002336502)
(25, 0.611590000140667)
(50, 0.6084600002944469)
(75, 0.5944900004267692)
  
        };
        \addlegendentry{GenAtt(0.530)}
        \addplot[mark=*,pink] plot coordinates {
           (0, 0.6122700000405311)
(5, 0.6035199997425079)
(10, 0.5953900006175041)
(15, 0.5917999997973442)
(20, 0.5817300004184246)
(25, 0.5750999996840954)
(50, 0.5228700000345707)
(75, 0.4462299995303154)
  
        };
        \addlegendentry{Rollout(0.462)}
        \addplot[mark=*,green] plot coordinates {
            (0, 0.6122700000405311)
(5, 0.6103900000095367)
(10, 0.6101800002455712)
(15, 0.6101200003385544)
(20, 0.6078200001239776)
(25, 0.6075800001621247)
(50, 0.5819100003480911)
(75, 0.48906999998092654)
  
        };
        \addlegendentry{RawAtt(0.496)}
    \addlegendentry{line}
    
    \end{axis}
    \end{tikzpicture} \\
    \begin{tikzpicture}[scale=0.4]
        \begin{axis}[
            xlabel=(c)\% of tokens removed,
            ylabel= accuracy,
            tick align=outside,
            legend style={fill=white, fill opacity=0.2, draw opacity=1, text opacity=1},
            legend pos=north east
            ]
        \addplot[mark=*,red] plot coordinates {
           (0, 0.6122700000405311)
(5, 0.6117700000166894)
(10, 0.5080600001096726)
(15, 0.43958999981880187)
(20, 0.383759999614954)
(25, 0.3306699989497662)
(50, 0.19119999969005586)
(75, 0.07932999985218048)
  
        };
        \addlegendentry{Ours(AUC=0.230)}
        \addplot[mark=*,blue] plot coordinates {
            (0, 0.6122700000405311)
(5, 0.611640000039339)
(10, 0.5022600003480912)
(15, 0.4262799995660782)
(20, 0.3834499994456768)
(25, 0.34505999977588653)
(50, 0.20407999926805495)
(75, 0.08402999983429908)
  
        };
        \addlegendentry{GenAtt(0.235)}
        \addplot[mark=*,pink] plot coordinates {
          (0, 0.6122700000405311)
(5, 0.6118400000333786)
(10, 0.5181100001513957)
(15, 0.45144000010490415)
(20, 0.41724999983906746)
(25, 0.3832599997997284)
(50, 0.23468999924659728)
(75, 0.08880999988913536)
  
        };
        \addlegendentry{Rollout(0.254)}
        \addplot[mark=*,green] plot coordinates {
            (0, 0.6122700000405311)
(5, 0.611640000039339)
(10, 0.5074000002980232)
(15, 0.4349800001621246)
(20, 0.3945799998223782)
(25, 0.35855999933481214)
(50, 0.22385999926924705)
(75, 0.08837999987602234)
  
        };
        \addlegendentry{RawAtt(0.245)}
    
    \end{axis}
    \end{tikzpicture}
    \begin{tikzpicture}[scale=0.4]
        \begin{axis}[
            xlabel=(d)\% of tokens removed,
            ylabel= accuracy,
            tick align=outside,
            legend style={fill=white, fill opacity=0.2, draw opacity=1, text opacity=1},
            legend pos=south west
            ]
        \addplot[mark=*,red] plot coordinates {
           (0, 0.6122700000405311)
(5, 0.6122700000405311)
(10, 0.612529999935627)
(15, 0.612350000011921)
(20, 0.61283000010252)
(25, 0.612150000333786)
(50, 0.5872100010335446)
(75, 0.48427999951839446)
  
        };
        \addlegendentry{Ours(AUC=0.498)}
        \addplot[mark=*,blue] plot coordinates {
  (0, 0.6122700000405311)
(5, 0.6122700000405311)
(10, 0.6116499999880791)
(15, 0.6114400001525879)
(20, 0.6110600003242492)
(25, 0.6086700001955032)
(50, 0.5566100001096725)
(75, 0.30098999923467634)
  
        };
        \addlegendentry{GenAtt(0.443)}
        \addplot[mark=*,pink] plot coordinates {
           (0, 0.6122700000405311)
(0, 0.6122700000405311)
(5, 0.6122700000405311)
(10, 0.5948000000119209)
(15, 0.5720099999785423)
(20, 0.5551300001978874)
(25, 0.5178899999439717)
(50, 0.2792899995803833)
(75, 0.04250999995470047)
  
        };
        \addlegendentry{Rollout(0.290)}
        \addplot[mark=*,green] plot coordinates {
         (0, 0.6122700000405311)
(5, 0.6122700000405311)
(10, 0.589079999935627)
(15, 0.55783999979496)
(20, 0.5374399998605252)
(25, 0.4966099996268749)
(50, 0.2398399994790554)
(75, 0.01700999994277954)
  
        };
        \addlegendentry{RawAtt(0.269)}
    \addlegendentry{line}
    
    \end{axis}
    \end{tikzpicture}
    \caption{Perturbation results of CLIPmapper-A. From left to right, they are (a) the positive perturbation on images, (b) the negative perturbation on images, (c) the positive perturbation on the texts, and (d) the negative perturbation on the texts.}
    \label{fig:perA}
\end{figure}
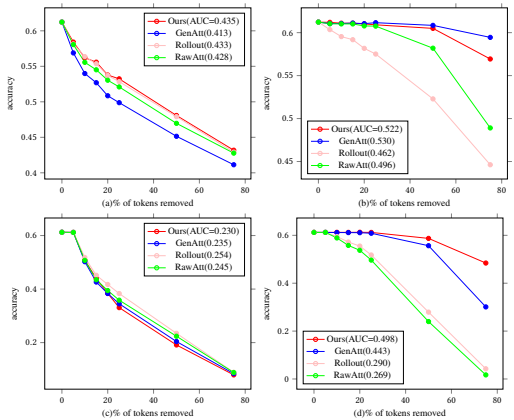
\subsubsection{Ours V.S. Others:} We separately conduct perturbation tests on images and texts of our reserving 10000 examples of VQA local testing. The scores are obtained from the row corresponding to the [CLS] token in scores matrices. We do perturbation tests both for CLIPmapper-A and CLIPmapper-B. The results are shown in Fig.~\ref{fig:perA} and \textbf{Appendix}.

It can be seen our method performs better than Rollout on perturbation tests of images, And our method outperforms GenAtt on perturbation tests of texts which means our method is effective.
\subsubsection{Validation Tests about the Explanation of the Whole Caption:} As Sec.~\ref{sec:eval} demonstrates, we choose BLEU4~\cite{bleu}, CIDEr~\cite{CIDEr}, and METEOR~\cite{METEOR} for weighting due to they are fast-computed methods. To evaluate whether the candidate caption matches the ground truth and the given image, we use RefCLIPScore~\cite{rCLIPs} that measure the matching degree among them. For comparison, we select the average as the baseline. We conduct both positive and negative tests on CLIPmapper-A.
 \begin{figure}[!htbp]
    \begin{tikzpicture}[scale=0.4]
        \begin{axis}[
            xlabel=(a)\% of tokens removed,
            ylabel= accuracy,
            tick align=outside,
            legend style={fill=white, fill opacity=0.2, draw opacity=1, text opacity=1},
            legend pos = north east
            ]
        \addplot[mark=*,green] plot coordinates {
            (0, 0.80365234375)
(5, 0.5263818359375)
(10, 0.51618408203125)
(15, 0.512001953125)
(20, 0.5150341796875)
(25, 0.51504638671875)
(50, 0.54501708984375)
(75, 0.5012548828125)
(100, 0.550556640625)
        };
        \addlegendentry{Average(AUC=0.531)}
        \addplot[mark=*,blue] plot coordinates {
            (0, 0.80365234375)
(5, 0.52771484375)
(10, 0.5158837890625)
(15, 0.52486572265625)
(20, 0.5125)
(25, 0.52224365234375)
(50, 0.52645263671875)
(75, 0.514208984375)
(100, 0.550556640625)
  
        };
        \addlegendentry{BLEU(0.531)}
        \addplot[mark=*,yellow] plot coordinates {
           (0, 0.80365234375)
(5, 0.52817138671875)
(10, 0.5175634765625)
(15, 0.52004638671875)
(20, 0.51216064453125)
(25, 0.5193212890625)
(50, 0.5416015625)
(75, 0.51739013671875)
(100, 0.550556640625)
  
        };
        \addlegendentry{CIDEr(0.535)}
        \addplot[mark=*,brown] plot coordinates {
            (0, 0.80365234375)
(5, 0.5305224609375)
(10, 0.5208251953125)
(15, 0.522998046875)
(20, 0.5177880859375)
(25, 0.51874755859375)
(50, 0.51593994140625)
(75, 0.48940673828125)
(100, 0.550556640625)
        };
        \addlegendentry{METEOR(0.523)}
    \addlegendentry{line}
    
    \end{axis}
    \end{tikzpicture}
    \begin{tikzpicture}[scale=0.4]
        \begin{axis}[
            xlabel=(a)\% of tokens removed,
            ylabel= accuracy,
            tick align=outside,
            legend style={fill=white, fill opacity=0.2, draw opacity=1, text opacity=1},
            legend pos = south west
            ]
        \addplot[mark=*,red] plot coordinates {
            (0, 0.80365234375)
(5, 0.8043994140625)
(10, 0.8039453125)
(15, 0.806171875)
(20, 0.8024951171875)
(25, 0.7981689453125)
(50, 0.7858837890625)
(75, 0.7348193359375)
(100, 0.550556640625)
  
        };
        \addlegendentry{Average(AUC=0.75)}
        \addplot[mark=*,blue] plot coordinates {
            (0, 0.80365234375)
(5, 0.78656494140625)
(10, 0.79114013671875)
(15, 0.790849609375)
(20, 0.78450439453125)
(25, 0.784296875)
(50, 0.77126708984375)
(75, 0.72055419921875)
(100, 0.550556640625)
  
        };
        \addlegendentry{BLEU(0.737)}
        \addplot[mark=*,yellow] plot coordinates {
           (0, 0.80365234375)
(5, 0.804755859375)
(10, 0.8047607421875)
(15, 0.8064794921875)
(20, 0.799423828125)
(25, 0.798544921875)
(50, 0.7827490234375)
(75, 0.72112548828125)
(100, 0.550556640625)
  
        };
        \addlegendentry{CIDEr(0.745)}
        \addplot[mark=*,brown] plot coordinates {
            (0, 0.80365234375)
(5, 0.5305224609375)
(10, 0.5208251953125)
(15, 0.522998046875)
(20, 0.5177880859375)
(25, 0.51874755859375)
(50, 0.51593994140625)
(75, 0.48940673828125)
(100, 0.550556640625)
  
        };
        \addlegendentry{METEOR(0.527)}
    \addlegendentry{line}
    
    \end{axis}
    \end{tikzpicture}
    \caption{The perturbation tests on CLIPmapper-B for the image captioning task.}
    \label{fig:perCap}
\end{figure}
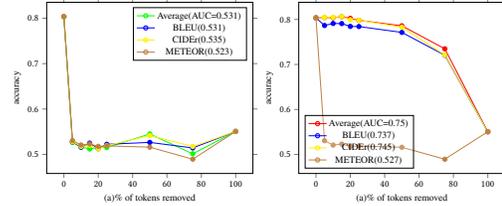

The results shown in Fig.~\ref{fig:perCap} are beyond our expectation in that it shows there are no significant differences between different weighting strategies to generate a faithful explanation of the whole caption.

\subsection{Explanation and Explainability Analysis}
\begin{figure}[!htbp]
	\begin{center}
		\includegraphics[width=0.66\linewidth]{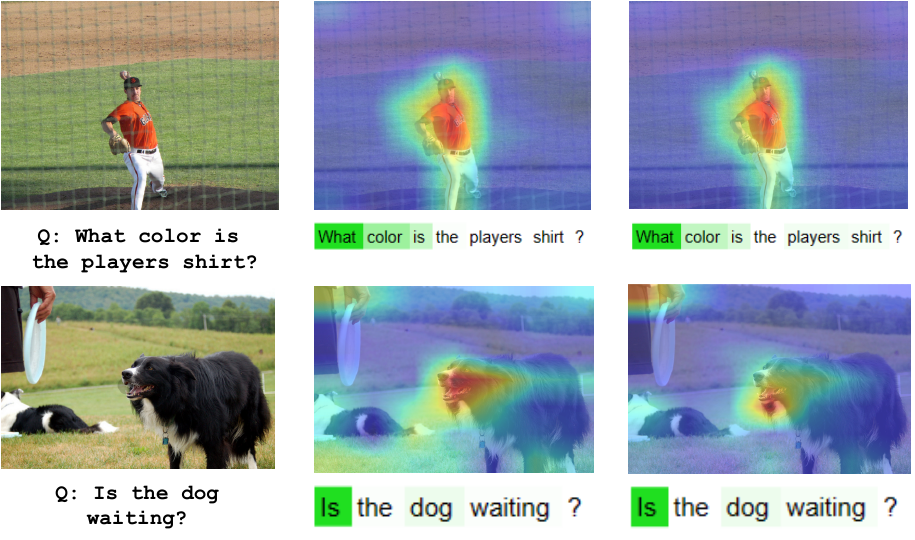}
	\end{center}
	\caption{The Visualization of Our Explanation on randomly selected 2 examples on VQA task. The middle column is the explanation of CLIPmapper-A and the right column is the explanation of CLIPmapper-B.}
	\label{fig:expVqa}
\end{figure}
We visualize explanations of inference outcomes of our CLIPmapper-A and CLIPmapper-B about randomly selected 4 examples in VQAv2. As can be seen in Fig.~\ref{fig:expVqa}, our explanation is reasonable. There are \textbf{more visualizations and discussions in the Appendix}.

\section{Conclusion}
In this paper, we implement the idea of Weighted Relevance Accumulation, which considers a better adaptive weighting to reduce the distortion of relevancy aggregation from the rigid average weighting of the existing method. Ablation study and perturbation tests indicate the effectiveness of our approach. Our work still has various limitations. Firstly, our approach sacrifices time complexity for accuracy. Besides, we cannot explain further complex generation tasks. However, these limitations point out the hopeful direction. We shall find out a better way to replace the InputGrad/AttGrad which both reduces the computational cost and extend the explanation to 
other generative tasks.

\clearpage
{
\bibliographystyle{ACM-Reference-Format}
\bibliography{all.bib}
}

\clearpage

\appendix
\balance
\section{Baseline Methods for Comparison}
We are concerned about the explainability of attention models. The current effective methods are Rollout~\cite{Rollout}, TransAtt~\cite{TransAtt}, GradCAM~\cite{gradCAM}, PRLP~\cite{PLRP} and GenAtt~\cite{GenAtt}. Among these methods, GenAtt has state-of-the-art performance and our methods can be seen as an improvement to Rollout so that we validate our method in the comparison with raw attention, Rollout, and GenAtt. We first initialize four score matrices as follows ($\mathbf{I, T}$ represent vision and language modalities, $\mathbf{i}$ and $\mathbf{t}$ respectively means token numbers of these two modalities):
\begin{align}
     &\label{eq24}\mathbf{R_{ii} = \mathbb{I}_{I\times I}, R_{tt} = \mathbb{I}_{T\times T}}\\
     &\label{eq25}\mathbf{R_{it} = {0}_{I\times T}, R_{ti} = {0}_{T\times I}}.
\end{align}
\textbf{Raw Attention:} Like usual attention visualization, they all take the attention of the last layer as an explanation. That is taking the last attention to updating corresponding score matrices only once.

\textbf{Rollout:} For the Rollout baseline, it is the same as the test settings of GenAtt. Rollout updates their self-modality score matrices Eq.~\ref{eq24} by multiplying all average attention across the self-attention head. Chefer et al. use the last cross-attention to update the cross-modality score matrices Eq.~\ref{eq25}. These rules are ( $\mathbf{\bar{A}_{-1}}$ is the last bi-modal attention):
\begin{align}
     &\label{eq26}\mathbf{R_{ss} = R_{SS}\cdot(\bar{A} + \mathbb{I})}\\
    &\label{eq27} \mathbf{\bar{A} = E_h(A)}\\
    &\label{eq28} \mathbf{R_{sq} = R_{sq}\cdot \bar{A}_{-1} \cdot R_{qq}}.
\end{align}
\textbf{GenAtt:} When meeting the self-attention of modality $\mathbf{S}$, GenAtt uses the following equation for updating:
\begin{align}
&\label{eq29} \mathbf{R_{ss} = R_{ss} + \bar{A} \cdot R_{ss}}\\
&\label{eq30} \mathbf{R_{sq} = R_{sq} + \bar{A}\cdot R_{sq}}\\
&\label{eq31} \mathbf{\bar{A} = E_{h}((\nabla A \odot A)^{+})}\\
&\label{eq32} \mathbf{\nabla A = \frac{\partial z_t}{\partial A}}.
\end{align}
When meeting the cross-attention that the query whose modality is $\mathbf{S}$ and the key/value whose modality is $\mathbf{Q}$, GenAtt uses the following equation for updating ($\mathbf{x}$ means token numbers of any modalities):
\begin{align}
    &\label{eq33}\mathbf{R_{sq} = R_{sq} + (\bar{R}_{ss})^T \cdot \bar{A} \cdot \bar{R}_{qq}}\\
  &\label{eq34}\mathbf{\hat{R}_{xx} = R_{xx} - \mathbb{I}}\\
  &\label{eq35}\mathbf{\bar{R} = norm_{row}(\hat{R}_{xx})+ \mathbb{I}}\\
  &\label{eq36}\mathbf{R_{ss} = R_{ss} + \bar{A}\cdot R_{qs}}.
\end{align}

\section{Results of Perturbation Test of CLIPmapper-B}

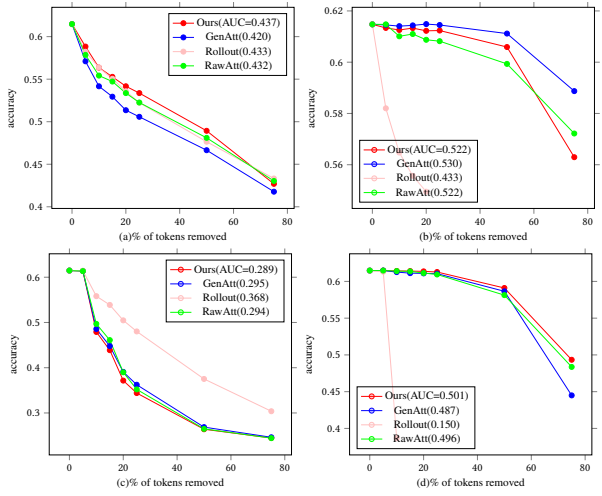
\begin{figure}[!ht]
    \begin{tikzpicture}[scale=0.47]
        \begin{axis} [
            xlabel=(a)\% of tokens removed,
            ylabel= accuracy,
            tick align=outside,
            legend pos = north east
        ]
        \addplot[mark=*,red] plot coordinates {
        (0, 0.6147600009441375)
(5, 0.5883200006008148)
(10, 0.5635800007104874)
(15, 0.5527300005674363)
(20, 0.5416700004339218)
(25, 0.5336500013828278)
(50, 0.4893100008964539)
(75, 0.42696000168323517)
        };
        \addlegendentry{Ours(AUC=0.437)}
        
        \addplot[mark=*,blue] plot coordinates {
            (0, 0.6147600009441375)
(5, 0.5710000012874603)
(10, 0.5416300007104874)
(15, 0.5294100015878678)
(20, 0.5135200001239777)
(25, 0.5057700003623963)
(50, 0.4664999994516373)
(75, 0.41775000009536745)
        };
        \addlegendentry{GenAtt(0.420)}
        
        \addplot[mark=*,pink] plot coordinates {
           (0, 0.6147600009441375)
(5, 0.5822500012874603)
(10, 0.5637100010871887)
(15, 0.5504500004768371)
(20, 0.5357300011157989)
(25, 0.522279999756813)
(50, 0.47648000166416166)
(75, 0.433250000667572)
        };
        \addlegendentry{Rollout(0.433)}
        
        \addplot[mark=*,green] plot coordinates {
(0, 0.6147600009441375)
(5, 0.5785600013256073)
(10, 0.55421000020504)
(15, 0.5472200001001358)
(20, 0.533649998998642)
(25, 0.5225600005865098)
(50, 0.48103000020980835)
(75, 0.4303399999141693)
        };
        \addlegendentry{RawAtt(0.432)}
    \label{ipa}
    \end{axis}
    \end{tikzpicture}
    \begin{tikzpicture}[scale=0.47]
        \begin{axis}[
            xlabel=(b)\% of tokens removed,
            ylabel= accuracy,
            tick align=outside,
            legend style={fill=white, fill opacity=0.2, draw opacity=1, text opacity=1},
            legend pos=south west
            ]
        \addplot[mark=*,red] plot coordinates {
            (0, 0.6147600009441375)
(5, 0.6133700007915497)
(10, 0.6125700010299683)
(15, 0.6133000009536743)
(20, 0.6122400013923645)
(25, 0.6123500011444092)
(50, 0.6059200006484985)
(75, 0.5629399991750718)
        };
        \addlegendentry{Ours(AUC=0.522)}
        
        \addplot[mark=*,blue] plot coordinates {
        (0, 0.6147600009441375)
(5, 0.6145400005817413)
(10, 0.6140700006484985)
(15, 0.6143600007534027)
(20, 0.6148700009822845)
(25, 0.6145100008010864)
(50, 0.61117000207901)
(75, 0.5887200006246567)
  
        };
        \addlegendentry{GenAtt(0.530)}
        
        \addplot[mark=*,pink] plot coordinates {
           (0, 0.6147600009441375)
(5, 0.5820100013971329)
(10, 0.5647400010824204)
(15, 0.5556700013399124)
(20, 0.549480001449585)
        };
        \addlegendentry{Rollout(0.433)}
        
        \addplot[mark=*,green] plot coordinates {
            (0, 0.6147600009441375)
(5, 0.6147600008010864)
(10, 0.6100999997615815)
(15, 0.6109900003194809)
(20, 0.6087400005102157)
(25, 0.6082000007152557)
(50, 0.5993400011301041)
(75, 0.5721800030231475)
        };
        \addlegendentry{RawAtt(0.522)}
   
    \end{axis}
    \end{tikzpicture} \\
    \begin{tikzpicture}[scale=0.47]
        \begin{axis}[
            xlabel=(c)\% of tokens removed,
            ylabel= accuracy,
            tick align=outside,
            legend style={fill=white, fill opacity=0.2, draw opacity=1, text opacity=1},
            legend pos=north east
            ]
        \addplot[mark=*,red] plot coordinates {
 (0, 0.6147600009441375)
(5, 0.6138300010204315)
(10, 0.4790500015616417)
(15, 0.43881000131368636)
(20, 0.37152000039219857)
(25, 0.34402000076770783)
(50, 0.26399000024199487)
(75, 0.24470999993681908)
        };
        \addlegendentry{Ours(AUC=0.289)}
        \addplot[mark=*,blue] plot coordinates {
            (0, 0.6147600009441375)
(5, 0.613950001001358)
(10, 0.4855400020837784)
(15, 0.44788000226020813)
(20, 0.3907500017106533)
(25, 0.3623700013458729)
(50, 0.26873000046014783)
(75, 0.2462699997127056)
        };
        \addlegendentry{GenAtt(0.295)}
        \addplot[mark=*,pink] plot coordinates {
           (0, 0.6147600009441375)
(5, 0.6143700010299683)
(10, 0.5581500008583069)
(15, 0.5387900019407272)
(20, 0.5046800026416779)
(25, 0.4801900018453598)
(50, 0.3751800012886524)
(75, 0.3037900001883507)
        };
        \addlegendentry{Rollout(0.368)}
        \addplot[mark=*,green] plot coordinates {
            (0, 0.6147600009441375)
(5, 0.6135100009441375)
(10, 0.4968500013589859)
(15, 0.4610500005722046)
(20, 0.3899200003743172)
(25, 0.3521000011444092)
(50, 0.2648800002515316)
(75, 0.2442999998807907)
        };
        \addlegendentry{RawAtt(0.294)}
   
    \end{axis}
    \end{tikzpicture}
    \begin{tikzpicture}[scale=0.47]
        \begin{axis}[
            xlabel=(d)\% of tokens removed,
            ylabel= accuracy,
            tick align=outside,
            legend style={fill=white, fill opacity=0.2, draw opacity=1, text opacity=1},
            legend pos=south west,
            ]
        \addplot[mark=*,red] plot coordinates {
            (0, 0.6147600009441375)
(5, 0.6147600009441375)
(10, 0.614220001077652)
(15, 0.6140700011730195)
(20, 0.6136500016212464)
(25, 0.6127900010585785)
(50, 0.5910500020503998)
(75, 0.49339000129699706)
  
        };
        \addlegendentry{Ours(AUC=0.501)}
        
        \addplot[mark=*,blue] plot coordinates {
            (0, 0.6147600009441375)
(5, 0.6147600009441375)
(10, 0.6126600011348724)
(15, 0.6112400012016297)
(20, 0.610910002040863)
(25, 0.6105400017261505)
(50, 0.5865400020122528)
(75, 0.4449400015830994)
        };
        \addlegendentry{GenAtt(0.487)}
        
        \addplot[mark=*,pink] plot coordinates {
          (0, 0.6147600009441375)
(5, 0.6132400009870529)
(10, 0.3867400001168251)
  
        };
        \addlegendentry{Rollout(0.150)}
        
        \addplot[mark=*,green] plot coordinates {
            (0, 0.6147600009441375)
(5, 0.6147600009441375)
(10, 0.6140500010967255)
(15, 0.6136500010967255)
(20, 0.6113400015830993)
(25, 0.6094200014591217)
(50, 0.5813400007724762)
(75, 0.4836499999642372)
        };
        \addlegendentry{RawAtt(0.496)}
    
    \end{axis}
    \end{tikzpicture}
    \caption{Perturbation tests' results of CLIPmapper-B. From left to right, they are (a) the positive perturbation on images, (b) the negative perturbation on images, (c) the positive perturbation on the texts, and (d) the negative perturbation on the texts.}
    \label{fig:perB}
\end{figure}

Results of Perturbation Test of CLIPmapper-B as shown in Fig.~\ref{fig:perB}. Our method has state-of-the-art performance on texts.

\section{Further Explanation and Explainability Analysis}
\subsection{More Visualization of Explanation on VQAv2}
\begin{figure}[b]
	\begin{center}
		\includegraphics[width=0.8\linewidth]{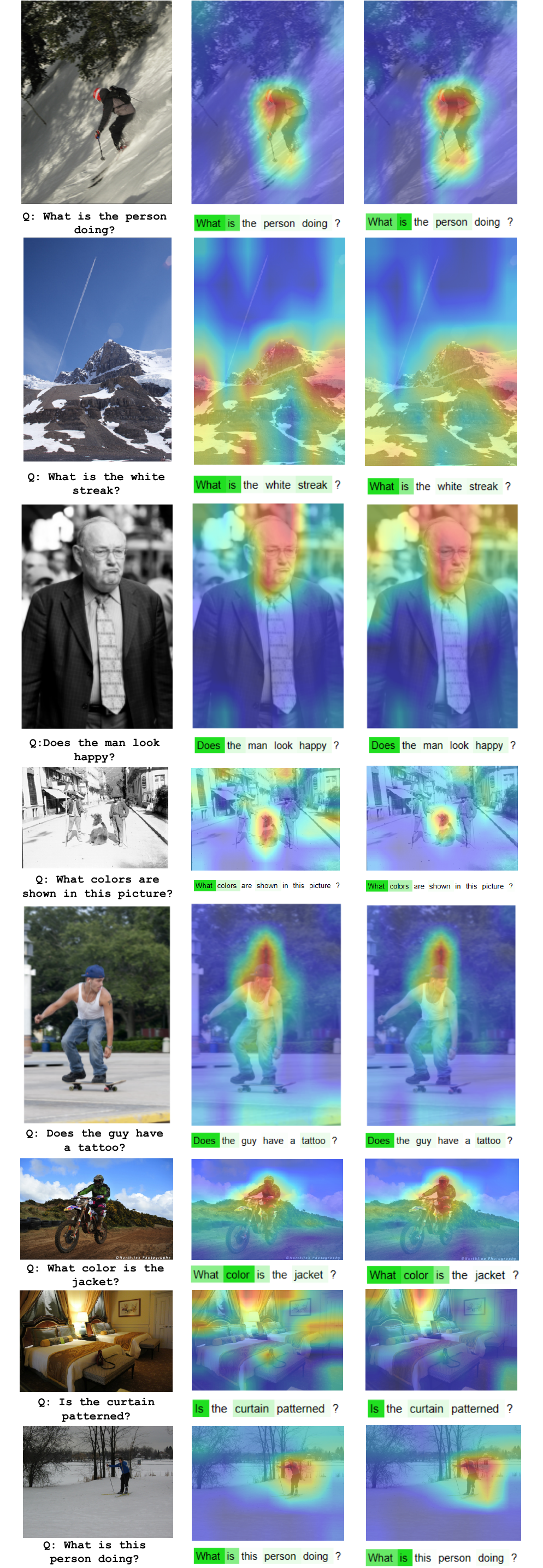}
	\end{center}
	\caption{The more visualization of our explanation on randomly selected 8 examples on VQA task. The middle column is the explanation of CLIPmapper-A and the right column is the explanation of CLIPmapper-B.}
	\label{fig:expVqa2}
\end{figure}

We provide more visualization of the explanation produced by our method on VQAv2 as Fig.~\ref{fig:expVqa2} shown.

\subsection{Visualization of Modality Interaction}

\begin{figure}[H]
	\begin{center}
		\includegraphics[width=0.7\linewidth]{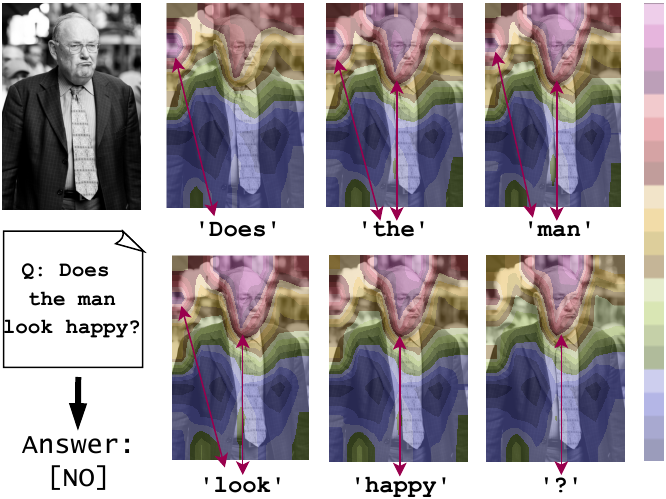}
	\end{center}
	\caption{The Visualization of Our Explanation on randomly selected 1 example on VQA task.}
	\label{fig:expInter}
\end{figure}

We visualize explanations of the modality interaction of CLIPmapper-A, which is computed by Eq.~\ref{eq23}. Fig.~\ref{fig:expInter} shows that the modality interaction related to the final output answer of the model is very homogeneous. 

\subsection{Averaging Explanation of Each Word}


Since the metric~\cite{rCLIPs} has a good characterization ability, we speculate that each word expresses too little meaning for the whole caption and only the chunk of words can represent a part of the meaning. In other words, it is considered to accept the average of concerns of each word about the image as the explanation of the caption\label{sec:why}. 

\subsection{Explanation of MSCOCO 2014 Caption Examples with Visualization}
\begin{figure}[H]
	\begin{center}
		\includegraphics[width=0.8\linewidth]{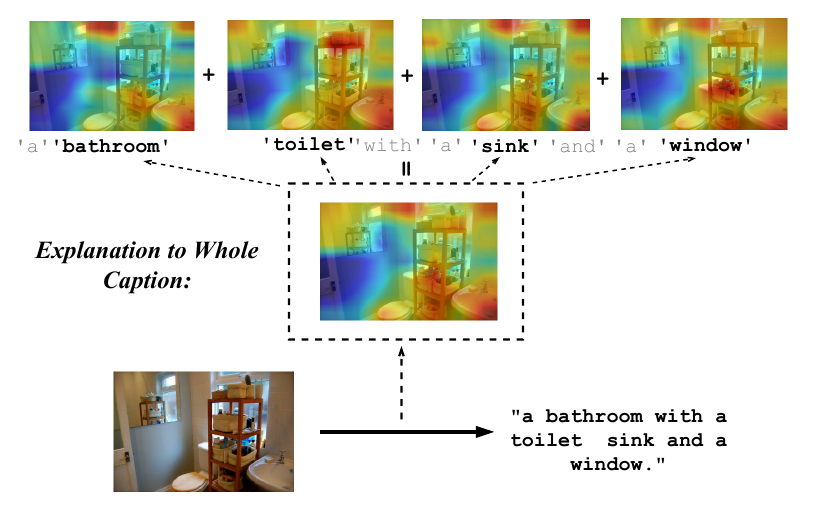}
	\end{center}
	\caption{The Visualization of Our Explanation on randomly selected 1 example on the image captioning task. It is produced in CLIPmapper-A. They all use meteor for weighting.}
	\label{fig:expCap}
\end{figure}
Like our analysis in Sec.~\ref{sec:why}, we choose meteor as the $\mathbf{eval(.)}$ of Eq.~\ref{eq19} and visualize our explanation as Fig.~\ref{fig:expCap} shows. 

\end{document}